# Look Inside. Predicting Stock Prices by Analysing an Enterprise Intranet Social Network and Using Word Co-Occurrence Networks

Fronzetti Colladon, A., & Scettri, G.

# Look Inside. Predicting Stock Prices by Analysing an Enterprise Intranet Social Network and Using Word Co-Occurrence Networks

Fronzetti Colladon, A., & Scettri, G.

**Abstract**

This study looks into the employees' communication behaviours taking place in an intranet social network, offering novel metrics of Semantic and Social Network Analysis, which can help predict a company stock price. To this purpose, we studied the intranet forum of a large Italian company, exploring the interactions and the use of language of about 8,000 employees. We analysed more than 48,000 news and comments, over a period of 94 weeks. In addition to using more traditional semantic and social network metrics, we built a network linking words included in the general discourse. In this network, we focused on the position of the node representing the company brand. We found that a lower sentiment of the language used, a higher betweenness centrality of the company brand, a denser word co-occurrence network and more equally distributed centrality scores of employees (lower group betweenness centrality) are all significant predictors of higher stock prices. Our findings contribute to the stream of research concerned with the prediction of stock prices, offering new metrics that can be helpful for scholars, company managers and professional investors and could be integrated into existing forecasting models to improve their accuracy. We also show the importance of looking at internal communication streams while analysing a company's financial performance. Lastly, we contribute to the research on word co-occurrence networks by extending their field of application.

**Keywords:** stock price; economic forecasting; intranet; social network; web forum; semantic analysis; word co-occurrence network.

## 1. Introduction

The question whether stock market prices are predictable is frequent in literature, especially after financial crisis (Cowles, 1933; Bollerslev and Ole Mikkelsen, 1996; Barro, 2015). In recent years, new approaches were presented, comprising the analysis of complex systems and data mining and machine learning techniques (Kuo, Chen and Hwang, 2001; Sornette, 2003). With the complex system approach it is possible to model the influence of the various agents that operate in a market (investors, companies, banks, nations, etc..), investigating their reciprocal interactions (Grimm, 2005). Data mining offers the possibility to analyse large volumes of data, retrieved from different sources, such as stock indices, social media or other online platforms. Nowadays researchers can take advantage of datasets never seen before, in which interesting information can be hidden.

In this paper, we combine methods and tools from Social Network and Semantic Analysis (Wasserman and Faust, 1994; Aggarwal and Zhai, 2013), while investigating the interactions taking place in the intranet forum of a large Italian company. We study the interaction patterns of the company employees with regard to their activity levels, network positions and use of language, while posting news and comments. Additionally, we propose a novel method to transform the general discourse into a network of words (Danowski, 2009), thus analysing the positions and fluctuations of the company brand in that network. Our scope is to understand if there are hidden associations between these patterns and the company stock price at specific time lags. In this regard, we offer a novel contribution since we use a new information source – which, to the extent of our knowledge, has never been used for the same exact purpose – and study original metrics obtained from the combination of semantic and social network analysis. In other words, the objective of this paper is to extract new variables from the analysis of the communication taking place in a large intranet social network, showing the value of these new predictors for the forecasting of stock prices: we investigate if it is possible to look at employees' communication behaviours to better predict the firm market value. This research also offers an additional contribution to increase the knowledge about possible uses of word co-occurrence networks.

**2. Network and Semantic Analysis for Stock Market Predictions**

The use of network theory to analyse stock market trends evolved around two main approaches: empirical and theoretical. The first is centred on the experimental ways by which networks are created, as in the work of Sun et al.(2015), who connected stock prices with other market attributes (e.g., trading volumes or net returns). The theoretical approach, on the other hand, is focused on the evaluation of the problem in a general form: given some mathematical conditions related to the network, the aim is to demonstrate the existence and uniqueness of a solution (Barucca *et al.*, 2016). Studies applying network theories were successfully implemented to analyse the behaviour of Brazilian (Tabak *et al.*, 2009), Chinese (Huang, Zhuang and Yao, 2009) and Indian (Pan and Sinha, 2007) stock markets. Tabak et al. (2009) tried to figure out if the Brazilian stock price returns presented a power law distribution. Their results showed that, for most of the study period, a power law distribution is not representative of the phenomenon. This suggests that the dynamics of stock prices may change abruptly, being more complicated than a power law distribution, especially when critical events occur. Huang and colleagues (2009) used a threshold method to build a social network considering the correlations among 1080 stocks in the Chinese market and their daily prices for about four years. Testing the topological stability of the network, they found it to be robust against random vertex failures, but fragile against targeted attacks; in this way they offered insights for risk and portfolio management. Similarly, Pan and Sinha (2007) analysed the cross-correlation matrix of stock prices fluctuations in the Indian National Stock Exchange. This market exhibited stronger correlations when compared to more mature markets, such as the New York

Stock Exchange. Pan and Sinha (2007) showed that the existence of an internal structure made of multiple connected groups is a possible indicator of market development.

Other authors addressed the problem of interdependence between stock markets (Morana and Beltratti, 2008; Sedik and Williams, 2011), suggesting the inclusion of control indices while making stock price predictions. In particular, Sedik and Williams (2011) used a GARCH model to study the influence of the volatility of U.S. and regional equity markets on the conditional volatility of stock prices in the Gulf Cooperation Council's market. Morana and Beltratti (2008), showed a link among the variances of the stock market indices of US, UK, Japan and Germany.

Recently, the improvement of sentiment analysis and other social media related research, offered new possibilities, combining stock market predictions with metrics extracted from online platforms, such as Twitter or Facebook (Zhang, Fuehres and Gloor, 2011; Chen and Lazer, 2013; Makrehchi, Shah and Liao, 2013). Measuring the sentiment of the conversations about a specific company, scholars proved the influence of positive and negative feelings on stock market prices (Chung and Liu, 2011; Elshendy *et al.*, 2017). To this purpose, Khadjeh et al. (2014) and Schumaker and Chen (2009) used sentiment analysis combined with linear regression models and support vector machines. In the former study, authors used the text of breaking financial news-headlines to forecast currency movements in the foreign exchange market. In the latter study, authors succeeded in partially predicting future stock prices twenty minutes after a financial article was released: they used several different textual representations – such as Bag of Words, Noun Phrases (Caropreso and Matwin, 2006) and Named Entities (Diesner and Carley, 2005). Zhang et al. (2011) analysed Twitter with a similar purpose, tagging tweets according to feelings of fear, general concern and hope; they found a negative correlation between the sentiment trend of the tagging variables and the Dow Jones, NASDAQ and S&P 500. Antweiler and Frank (2004) studied the influence of messages in Yahoo!Finance over a set of 45 stock market prices of private companies. Similarly, Xie et al. (2013) crawled Yahoo!Finance and developed a tree representation for words in sentences, which performed significantly better than approaches based on bag-of-words in predicting the polarity of stock prices trends. A different approach for the sentiment analysis of tweets is to use non-parametric topic modelling algorithms, as proposed by Si et al. (2013).

In the paper of Bollen Mao and Zeng (2011) the prediction of the Dow Jones Industrial Average was done by analysing the text content of daily Twitter feeds by means of two mood-tracking tools: OpinionFinder – which tracks positive and negative moods – and Google Profile of Mood States (GPOMS), which measures language mood in term of six dimensions (Calm, Alert, Sure, Vital, Kind, and Happy). Rechenthin et al. (2013) tried to understand if there were financial agents that could manipulate sentiment trends, to influence stock market prices posting specific, either positive or negative, messages on specialized financial websites. Nguyen et al. (2014) analyzed the main topics and sentiment of eighteen Yahoo!Finance message boards for eighteen stocks. On message boards, users discussed company news, facts or comments (often negative) about specific company events, and personal forecasts. Analysing such platforms can be helpful, also because users have the possibility to annotate message with tags (e.g., Strong Buy, Buy, Hold, Sell and Strong Sell). Authors proved that adding sentiment analysis to models based on historical prices trends can increase the predictive power of such models.

Yang et al. (2015) demonstrated the existence of Twitter's financial communities – which presented a small-world structure – inferred studying friend-following relationships and user profiles, including language preferences, locations, account creation dates and time zones. Looking at the sentiment of the tweets sent by people in the most central network positions, the authors could predict financial market indices.

Lastly, other techniques were also implemented with the aim of improving stock prices predictions, such as artificial neural networks. Patel and Yalamalle (2014), for example, achieved good results using neural networks to predict stock prices in India.

Semantic and social network analysis of the communication of employees proved their value for several purposes, such as predicting turnover intentions (Gloor et al., 2017a), improving customer satisfaction (Gloor and Giacomelli, 2014; Gloor et al., 2017b), or promoting innovation within, and across, organisational boundaries (Wright and Dana, 2003; Dana, Etemad and Wright, 2008; Allen et al., 2016). These analyses, which also stressed the importance of looking at the communication styles and interactions taking place within the organizational boundaries, were often carried out using traditional surveys or exploring e-mail networks.

Starting from the insights that emerge from past research, we developed the present study with the idea of offering a triple contribution: firstly, we present semantic and social network metrics that can be integrated in existing financial models to increase their forecasting accuracy; secondly, we give evidence to the value of transforming text data into words of networks, to extend the informative power of traditional semantic analysis; thirdly, we show how the market value of a firm can be at least partially inferred by looking at the internal interactions among employees, when considering an intranet social network.

An intranet is a private network based on web protocols, belonging to an organization, and usually accessible only by the organization's members. The websites and software applications of an intranet look and act just like any others, but the firewall surrounding an intranet fends off unauthorized access and use (Beal, 2017). An intranet forum is a social network among all the people inside a company, in which employees can exchange text messages, share news and comments, or multimedia files.

Enterprise intranets proved to offer important insights to business mangers (Eppler, 2001), also being a valuable tool to promote knowledge sharing (Hollingshead, Fulk and Monge, 2002), facilitate HR activities and assess the internal mood (Sulaiman, Zailani and Ramayah, 2012). From an intranet social network it is often possible to extract knowledge maps (Eppler, 2001). These are graphs that provide a visual information about knowledge sources and help evaluating strengths and weaknesses of knowledge related assets. Intranets are useful also from a Human Resource point of view, to understand roles and skills within the organization, current activities and possessed knowledge (DiMicco et al., 2009).

To the extent of our knowledge, there is no research trying to combine words of networks with semantic and social network analysis of an intranet social network, with the aim of forecasting the stock price of a company.

### 2.1. Words and Networks

In this paper, we propose the analysis of a word co-occurrence network, as an addition to traditional semantic analysis. There are several studies in this field, for example Diesner and Carley (2005) showed how to detect social structures through text analysis: by analysing major newspapers they managed to discover the social structure of covert networks – terrorist groups operating in the West Bank.

Networks of words can be built in different ways, such as the analysis of words co-occurrences in single sentences or text excerpts (Dagan, Marcus and Markovitch, 1995; Bullinaria and Levy, 2012). Centering resonance analysis was also proposed as a method for creating a network from a text by analysing its centres (Corman *et al.*, 2002). Other scholars used techniques based on hypertexts (Trigg and Weiser, 1986) or semantic webs (Tim and Berners-Lee, 2006; van Atteveldt, 2008) focusing on specific words and links between online pages.

In our explorative study, we created a word co-occurrence network, considering the messages extracted from the intranet forum of a large Italian company and the co-occurrence of these words. We investigated the position in the graph of the company brand – i.e. its centrality measures (Freeman, 1978) – to better predict the company stock price (together with the use of more conventional social network and semantic variables).

### 3. Case Study

In our case study, we were able to fully crawl the intranet forum of a large Italian company, with more than 50,000 employees, out of which about 8,000 were actively participating to the online discussions. As per agreed privacy arrangements, we are prohibited from revealing more details about the company.

In the intranet forum, employees were allowed to post news and to comment on their own posts or those of others, with no restrictions or pre-approvals from moderators. Starting from the analysis of these interaction patterns, we were able to create a first *Interaction Network*, with $n$ nodes (employees) and $m$ directed arcs (posts). In this network, there is an arc originating at the node $i$ and terminating at the node $j$, if the social actor $i$ answers to a comment or a news of the actor $j$. This network – built considering the news and comments posted between September 2014 and June 2016, for a total of 94 weeks – comprises 8320 nodes and 48020 links.

In addition to studying the interactions among people working in the company, we also analysed the relationships among the words used in the posts. In particular, we transformed the text of news and comments into a *Word Network*, based on words co-occurrences (Dagan, Marcus and Markovitch, 1995; Bullinaria and Levy, 2012). In this network, nodes are representative of single words and arcs connect words that co-occur in the text, either considering the words before, or after a specific one, with a maximum distance of seven words. To put it in other words, for each co-occurrence within this range we created a link going from one word to the other, with a directionality which respects the order of appearance in the text. To give an example, if a post is "Hello Dolly", then we would have two nodes – "Hello" and "Dolly" – with an arc originating at the first node and terminating at the second one. Before building the word network, we corrected or removed the misspelled words and removed the stop words - i.e. most common words in the language which usually do not contain important significance - using the Python NLTK library (Perkins, 2014). The resulting network is made of about 16,000 words and more than 6,000,000 links.

When building networks of words, different techniques can be used and the maximum range for considering a co-occurrence can vary. The choice of a maximum interval of 7 words provided the best results in our case study; however, we maintain that the analyst should be free to adjust this interval according to the specific context and dataset analysed. We also considered the possibility to simplify our network, by applying lemming or stemming algorithms to reduce words to their root forms (Korenius *et al.*, 2004; Perkins, 2014). However, a test in this direction produced no better results than those obtained when we worked on the full network.

### 3.1. Study Variables

Considering the two networks described in the previous section, we extracted weekly measures for a specific set of variables, representative of the social structure, the activity and the employees' use of language. To measure structural positions, we referred to well-known centrality metrics of Social Network Analysis: betweenness and degree centrality (Freeman, 1978).

*Degree centrality* measures the number of direct connection of a node in the network and corresponds to the number of incident arcs, either originating or terminating at that node.

*Group Degree Centrality* quantifies the variability of individual degree centrality scores and it is used to measure how much centralized a network is, and so how much dominated by a set of highly central nodes. This measure reaches its maximum of 1 for a star graph (Wasserman and Faust, 1994).

*Betweenness Centrality* is a measure that reflects the number of times a node lies in-between the shortest paths that connect every other pair of nodes (Wasserman and Faust, 1994).

*Group Betweenness Centrality* reflects the heterogeneity of betweenness centrality scores of single actors: for a star graph the value is maximum and equal to 1; for a network where each actor has the same level of betweenness, this index is equal to zero (Wasserman and Faust, 1994).

With regard to the interaction network, we considered group level metrics with the intent of investigating the full network activity, without focusing on single employees; for the word network, on the other hand, we focused on the structural positions of the node associated to the company brand (which also corresponds to the company name in the stock market). Our aim was to test if the fluctuations of the brand in this network could be used as a predictor of the company stock price at the end of each week (which is our dependent variable).

To investigate the use of language by the employees in the intranet forum, we carried out a semantic analysis of the content of news and comments, mapping their average weekly level of sentiment, emotionality and complexity.

*Sentiment* is a measure of the positivity or negativity of a text, calculated by using a multi-lingual classifier based on a machine learning algorithm (trained on large datasets extracted from Twitter) (Brönnimann, 2014). A score of 0.5 represents a neutral sentiment, whereas higher scores represent a positive sentiment, and lower scores a negative one.

*Emotionality* measures the variation in the overall sentiment and it is calculated as its standard deviation. If all the messages converge towards a positive sentiment score, the emotionality will be low; by contrast, if there is a frequent alternation of positive and negative messages, the emotionality will be high (Brönnimann, 2014).

*Complexity* measures the distribution of words within a text, considering as more complex the words that are less frequently used in a specific context (Brönnimann, 2014). Accordingly, a rare word that is commonly used in an intranet forum will not increase the general level of complexity.

As the last predictor in our study, we considered the weekly levels of activity as the number of messages posted during a week (*Activity*) or the number of word co-occurrences (*Activity Words*), i.e. the number of links in the word network.

The semantic analysis, as well as the computation of the centrality and activity measures, were implemented by using the software Condor[1] and the package NLTK developed for the programming language Python 3 (Perkins, 2014).

---

[1] www.galaxyadvisors.com/product-extended.html

Lastly, to control for the general stock market fluctuations, and the events which could affect all the market, we included the FTSE MIB in the predictive models. This index is the primary benchmark for the Italian equity market; it captures about 76% of the domestic market capitalization.

As per agreed privacy arrangements, we could not collect other control variables, such as employees' age, gender, job role or business department.

## 4. Results

We combined the variables presented in Section 3 to see if they could be used as predictors of a company price on the Italian stock market. Our aim is not to present a final model to make stock market predictions, but to draw the attention of scholars on new variables that can be extracted from a company intranet forum and easily integrated in more complex models, to improve their predictive power.

We started our analysis by looking at the correlations between our variables and the company stock price, at various time lags. Specifically, we analysed the insights coming from the intranet forum during the same week in which the stock price was collected (Current Week) and in the two weeks before (named respectively 1 Week Lag and 2 Weeks Lag), to see which time interval was more informative (Table 1).

|  | **Current Week Price** | **Price 1 Week Lag** | **Price 2 Weeks Lag** |
|---|---|---|---|
| Activity Words | .421** | .447** | .426** |
| Activity (Interaction Network) | -.128 | -.121 | -.180 |
| Group Betweenness Centrality (Interaction Network) | -.302** | -.256* | -.272** |
| Betweenness Centrality (Word Network) | .142 | .107 | .283* |
| Complexity (Interaction Network) | .091 | .043 | -.034 |
| Degree Centrality (Word Network) | .369** | .358** | .330** |
| Emotionality (Interaction Network) | -.165 | -.194 | -.148 |
| Sentiment (Interaction Network) | -.185 | -.230* | -.264* |
| FTSE MIB | .877** | .863** | .877** |
| Group Degree Centrality (Interaction Network) | -.389** | -.286** | -.252* |

*p<.05; **p<.01.

**Table 1.** Pearson's correlation coefficients with stock price (N=94).

For an easier reading, we only reported the correlation coefficients of each variable with the company stock price. As the table shows, there are relatively high correlations of group degree and group betweenness centrality with price. These correlations are negative at all lags, suggesting that a more heterogeneous network, less centred on few dominant nodes, is a better indicator of higher prices. As regards the semantic variables, sentiment is the only one which shows a significant correlation (at lag 1 and 2); this correlation is negative, probably reflecting the fact that a more technical language is usually associated to news related to firm performance and to financial events. As expected, FTSE MIB is highly correlated with stock price, proving its value as a control variable. Lastly, the degree centrality of the company brand in the word network proved to be significantly correlated with price: when the company name is more frequently mentioned in the general discourse, in association with a higher number of different co-occurring words, the stock price is higher. To consider possible autocorrelation effects, we implemented a first differencing of the dependent variable and the Granger causality tests presented in Table 2. We considered up to three-week lags, as suggested by the models diagnostics.

| Dependent: Price | $\chi^2$ |
|---|---|
| FTSE MIB | 3.796^ |
| Activity (Interaction Network) | .994^ |
| Activity Words | 5.246^ |
| Group Betweenness Centrality (Interaction Network) | 3.984^ |
| Betweenness Centrality (Word Network) | 3.777^ |
| Complexity (Interaction Network) | 5.806^ |
| Degree Centrality (Word Network) | .023^ |
| Emotionality (Interaction Network) | 1.543^ |
| Sentiment (Interaction Network) | .831^ |
| Group Degree Centrality (Interaction Network) | 1.574^ |

^$p > .05$.

**Table 2** Granger causality tests (Chi-squared values, N=94).

As Table 2 shows, all our predictors can significantly granger-cause the company stock price, suggesting that they could all be considered while making new predictive models.

As a last step of our analysis, we built multiple regression models to forecast the company stock price and comment on the variance explained by each predictor (Table 3). We did not use more complex time series models – such as ARIMAX – because the autocorrelation of the stock price was not significant in the Durbin-Watson test.

|  | Model 1 | Model 2 | Model 3 | Model 4 | Model 5 | Model 6 | Model 7 | Model 8 |
|---|---|---|---|---|---|---|---|---|
| FTSE MIB | 5.34x10E-5** |  |  |  |  |  |  | 5.88x10E-5** |
| Complexity (Interaction Network) Lag 0 |  | .029 |  |  |  |  |  |  |
| Emotionality (Interaction Network) Lag1 |  | -.058 |  |  |  |  |  |  |
| Sentiment (Interaction Network) Lag2 |  | -.108* |  |  |  |  |  | -.053** |
| Activity Words Lag1 |  |  | 0.222** |  |  |  |  | .076** |
| Activity (Interaction Network) Lag 0 |  |  |  | 1.59x10E-5 |  |  |  |  |
| Group Degree Centrality (Interaction Network) Lag 0 |  |  |  |  | -.409* |  |  |  |
| Group Betweenness Centrality (Interaction Network) Lag2 |  |  |  | -.236* |  |  |  | -.053* |
| Betweenness Centrality (Word Network) Lag 0 |  |  |  |  |  | .100* |  | .114** |
| Degree Centrality (Word Network) Lag0 |  |  |  |  |  |  | .001** |  |
| Constant | -.085 | 1.081** | .899** | 1.081** | 1.069* | .963** | .922** | -.179** |
| Adjusted R Squared | .758 | .046 | .205 | .069 | .142 | .032 | .127 | .851 |

*p<.05;
**p<.01

**Table 3.** Stock price prediction: multiple regression models.

In the first models, we tested the contribution of predictors in blocks, avoiding to put together measures for which we found multicollinearity problems (e.g., group degree and betweenness centrality). In Model 8, we combined all the significant variables which could better explain the price variance. With respect to the time lags of each variable, we chose those who better performed in the models. We found that the FTSE MIB alone explained 75.8% of the price variance. The new predictors introduced by this study could increase the variance explained by 9.2%. In particular, we found that negative sentiment at two weeks lag, higher activity in the word network at one week lag, higher betweenness centrality of the company brand and lower group betweenness centrality are all significant indicators of a higher stock price. Figure 1 summarizes our findings.

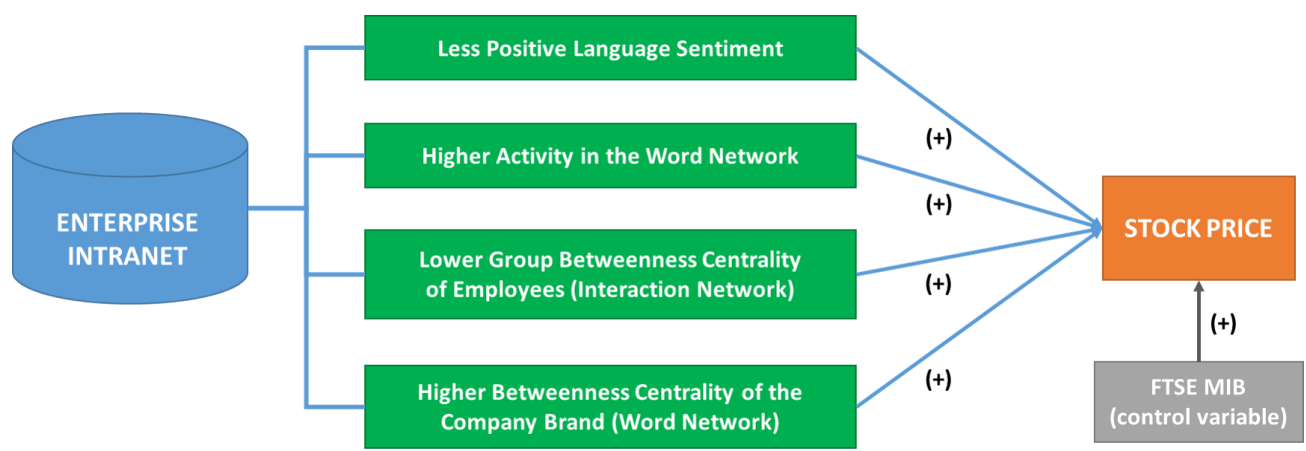

**Figure 1.** Significant predictors from the intranet social network.

## 5. Discussion and Conclusions

The objective of this paper was to study the employees' communication behaviour on a large company intranet forum, to extract new variables that could be useful to forecast a company's stock price. We did not mean to present final forecasting models but to give evidence to the importance of new metrics that can be integrated in existing financial models to improve their accuracy. In this sense, our work has practical and theoretical implications for scholars, financial analysts and professional investors.

By means of Granger causality tests, we proved the value of our metrics up to two-week lags. Subsequently, we included our predictors in multiple regression models to see whether they could improve the models accuracy with respect to the control variable FTSE MIB.

Our work is consistent with previous studies linking sentiment analysis and stock prices (Chung and Liu, 2011; Zhang, Fuehres and Gloor, 2011): we found a negative correlation of the stock price with the sentiment of messages posted in the company intranet forum. This could be partially explained by a more technical language used in news which deal with firm performance (with a lower sentiment, for example, when compared to posts where employees discuss their leisure activities promoted by the company). It could be the case that more irrelevant and positive messages are dominant in the network when there are no big events that can positively affect the stock price (this hypothesis was partially confirmed by an interview with the people in charge of monitoring the employees' activity on the intranet forum). A more "democratic" network structure (lower group betweenness centrality), less dominated by a smaller number of social actors, also proved to be predictive of higher prices.

We offer a new contribution since we are among the first scholars linking a company stock price with the communication behaviours on its intranet forum. Even if the informative power of the analysis of the employees' activity on enterprise intranets has been discussed in the past (Eppler, 2001; DiMicco et al., 2009), its association with the market value of a company remains mostly unexplored. We support the idea that internal communication behaviours can affect and partially mirror business performance (Gloor et al., 2017b), thus influencing a company's stock price.

An additional contribution of this research is to present a new use of word co-occurrence networks, which helps financial predictions. We transformed the content of the messages posted on the intranet forum into a word co-occurrence network – where every node, representative of a specific word, was linked to the others based on their co-occurrences in messages (Dagan, Marcus and Markovitch, 1995); on this network, we studied the associations between specific positions of the company brand and its stock price. Looking at the patterns interconnecting words, we saw that a higher activity – i.e. a denser graph with more co-occurrences – and a higher betweenness centrality of the company brand can predict higher prices. This last result could be representative of cases where the brand is more widely used, spreading across many different news and comments and therefore being more central in the discourse.

Some of the limitations of this study are due to our impossibility to analyse more companies or different business sectors. Moreover, we could not consider the communication and the interactions between employees which went through other media – such as emails, instant messaging apps, phone calls or face-to-face communication.

To extend our results, we advocate further research to study the intranet forums of more firms, also taking into account their size and business sector. It would also be important to consider larger time frames, to test other variables and network construction techniques, and to develop predictive models based on more complex machine learning algorithms (which already proved their value in the past).

Due to privacy agreements, we could not access to more control variables to differentiate the interactions based on the individual differences of employees. Therefore, future researchers should consider taking into account factors such as employees' age, gender and job role. Lastly, it could be interesting to study if more accurate models can be obtained by isolating the communication taking place in specific business departments or geographical locations. Depending on the business context, one could focus the attention to the communication of employees working in specific business units (customer care, financial department, etc...).

## References

Aggarwal, C. C. and Zhai, C. X. (2013) *Mining text data*, *Mining Text Data*. Edited by C. C. Aggarwal and C. Zhai. Boston, MA: Springer US. doi: 10.1007/978-1-4614-3223-4.

Allen, T. J., Gloor, P. A., Fronzetti Colladon, A., Woerner, S. L. and Raz, O. (2016) 'The Power of Reciprocal Knowledge Sharing Relationships for Startup Success', *Journal of Small Business and Enterprise Development*, 23(3), pp. 636–651. doi: 10.1108/JSBED-08-2015-0110.

Antweiler, W. and Frank, M. Z. (2004) 'Is All That Talk Just Noise? The Information Content of Internet Stock Message Boards', *The Journal of Finance*, 59(3), pp. 1259–1294. doi: 10.1111/j.1540-6261.2004.00662.x.

van Atteveldt, W. H. (2008) *Semantic Network Analysis Techniques for Extracting , Representing ,*

*and Querying Media Content*. Charleston, SC: BookSurge Publishing.

Barro, R. J. (2015) 'The Stock Market and Investment', *The Review of Financial Studies*, 3(1), p. 115. doi: 10.1093/rfs/3.1.115.

Barucca, P., Bardoscia, M., Caccioli, F., D'Errico, M., Visentin, G., Battiston, S. and Caldarelli, G. (2016) 'Network Valuation in Financial Systems', available at SSRN: https://ssrn.com/abstract=2795583

Beal, V. (2017) *Intranet, Webopedia*. Available at: http://www.webopedia.com/TERM/I/Intranet.html (Accessed: 23 May 2017).Bollen, J., Mao, H. and Zeng, X. (2011) 'Twitter mood predicts the stock market', *Journal of Computational Science*, 2(1), pp. 1–8. doi: 10.1016/j.jocs.2010.12.007.

Bollerslev, T. and Ole Mikkelsen, H. (1996) 'Modeling and pricing long memory in stock market volatility', *Journal of Econometrics*, 73(1), pp. 151–184. doi: 10.1016/0304-4076(95)01736-4.

Brönnimann, L. (2014) *Analyse der Verbreitung von Innovationen in sozialen Netzwerken*. University of Applied Sciences Northwestern Switzerland.

Bullinaria, J. a and Levy, J. P. (2012) 'Extracting semantic representations from word co-occurrence statistics: stop-lists, stemming, and SVD.', *Behavior research methods*, 44(3), pp. 890–907. doi: 10.3758/s13428-011-0183-8.

Caropreso, M. F. and Matwin, S. (2006) 'Beyond the Bag of Words: A Text Representation for Sentence Selection', in *Lecture Notes in Computer Science (including subseries Lecture Notes in Artificial Intelligence and Lecture Notes in Bioinformatics)*, pp. 324–335. doi: 10.1007/11766247_28.

Chen, R. and Lazer, M. (2013) *Sentiment Analysis of Twitter Feeds for the Prediction of Stock Market Movement*, Cs 229 Machine Learning: Final Project, University of Stanford.

Chung, S. and Liu, S. (2011) *Predicting stock market fluctuations from Twitter: A analysis of the predictive powers of real-time social media*. University of Berkeley, Berkeley, CA.

Corman, S. R., Kuhn, T., Mcphee, R. D. and Dooley, K. J. (2002) 'Studying Complex Discursive Systems Centering Resonance Analysis of Communication', *Human Communication Research*, pp. 157–206. doi: 10.1093/hcr/28.2.157.

Cowles, A. (1933) 'Can Stock Market Forecasters Forecast?', *Econometrica*, 1(3), pp. 309–324. doi: 10.2307/1907042.

Dagan, I., Marcus, S. and Markovitch, S. (1995) 'Contextual Word Similarity and Estimation from Sparse Data', *Computer Speech and Language*, 9, pp. 123–152.

Dana, L. P., Etemad, H. and Wright, R. W. (2008) 'Toward a paradigm of symbiotic entrepreneurship', *International Journal of Entrepreneurship and Small Business*, 5(2), pp. 109–126. doi: 10.1504/IJESB.2008.016587.

Danowski, J. (2009) 'Inferences from word networks in messages', in Krippendorff, K. and Bock, M. A. (eds) *The content analysis reader*. London, UK: Sage Publications, Inc, pp. 421–429.

Diesner, J. and Carley, K. M. (2005) 'Revealing Social Structure from Texts', in *Causal Mapping for Research in Information Technology*. IGI Global, pp. 81–108. doi: 10.4018/978-1-59140-396-8.ch004.

DiMicco, J. M., Geyer, W., Millen, D. R., Dugan, C. and Brownholtz, B. (2009) 'People Sensemaking and Relationship Building on an Enterprise Social Network Site', in *2009 42nd Hawaii International Conference on System Sciences*. IEEE, pp. 1–10. doi: 10.1109/HICSS.2009.343.

Elshendy, M., Fronzetti Colladon, A., Battistoni, E. and Gloor, P. A. (2017) 'Using Four Different Online Media Sources to Forecast the Crude Oil Price', *Journal of Information Science*, p. in press. doi: 10.1177/0165551517698298

Eppler, M. J. (2001) 'Making knowledge visible through intranet knowledge maps: concepts, elements, cases', in *Proceedings of the 34th Annual Hawaii International Conference on System Sciences*. IEEE Comput. Soc. doi: 10.1109/HICSS.2001.926495.

Freeman, L. C. (1978) 'Centrality in Social Networks', *Social Networks*, 1(3), pp. 215–239. doi:

10.1016/0378-8733(78)90021-7.

Gloor, P. A., Fronzetti Colladon, A., Grippa, F. and Giacomelli, G. (2017a) 'Forecasting managerial turnover through e-mail based social network analysis', *Computers in Human Behavior*, 71, pp. 343–352. doi: 10.1016/j.chb.2017.02.017.

Gloor, P. A., Fronzetti Colladon, A., Grippa, F., Giacomelli, G., and Saran, T. (2017b) 'The Impact of Virtual Mirroring on Customer Satisfaction', *Journal of Business Research*, 75, pp. 67-76. doi: 10.1016/j.jbusres.2017.02.010

Gloor, P. A. and Giacomelli, G. (2014) 'Reading Global Clients' Signals', *MIT Sloan Management Review*, 55(3), pp. 23–29.

Grimm, V. (2005) 'Pattern-Oriented Modeling of Agent-Based Complex Systems: Lessons from Ecology', *Science*, 310(5750), pp. 987–991. doi: 10.1126/science.1116681.

Hollingshead, A. B., Fulk, J. and Monge, P. (2002) 'Fostering Intranet Knowledge Sharing: An Integration of Transactive Memory and Public Goods Approaches BT - Distributed Work', in *Distributed Work*, pp. 335–355.

Huang, W., Zhuang, X. and Yao, S. (2009) 'A network analysis of the Chinese stock market', *Physica A,* 388(14), pp. 2956–2964. doi: 10.1016/j.physa.2009.03.028.

Khadjeh Nassirtoussi, A., Aghabozorgi, S., Ying Wah, T. and Ngo, D. C. L. (2014) 'Text mining for market prediction: A systematic review', *Expert Systems with Applications*. Elsevier Ltd, 41(16), pp. 7653–7670. doi: 10.1016/j.eswa.2014.06.009.

Korenius, T., Laurikkala, J., Järvelin, K. and Juhola, M. (2004) 'Stemming and lemmatization in the clustering of finnish text documents', in *Proceedings of the Thirteenth ACM conference on Information and knowledge management - CIKM '04*. New York, NY: ACM Press, pp. 625-633. doi: 10.1145/1031171.1031285.

Kuo, R. J., Chen, C. H. and Hwang, Y. C. (2001) 'An intelligent stock trading decision support system through integration of genetic algorithm based fuzzy neural network and artificial neural network', *Fuzzy Sets and Systems*, 118(1), pp. 21–45. doi: 10.1016/S0165-

0114(98)00399-6.

Makrehchi, M., Shah, S. and Liao, W. (2013) 'Stock Prediction Using Event-Based Sentiment Analysis', in *2013 IEEE/WIC/ACM International Joint Conferences on Web Intelligence (WI) and Intelligent Agent Technologies (IAT)*. IEEE, pp. 337–342. doi: 10.1109/WI-IAT.2013.48.

Morana, C. and Beltratti, A. (2008) 'Comovements in international stock markets', *Journal of International Financial Markets, Institutions and Money*, 18(1), pp. 31–45. doi: 10.1016/j.intfin.2006.05.001.

Nguyen, T. H., Shirai, K. and Velcin, J. (2014) 'Sentiment analysis on social media for stock movement prediction', *Expert Systems with Applications*. Elsevier Ltd., 42(24), pp. 9603–9611. doi: 10.1016/j.eswa.2015.07.052.

Pan, R. K. and Sinha, S. (2007) 'Collective behavior of stock price movements in an emerging market', *Physical Review E*, 76(046116), pp. 1-9. doi: 10.1103/PhysRevE.76.046116.

Patel, M. B. and Yalamalle, S. R. (2014) 'Stock Price Prediction Using Artificial Neural Network', *International Journal of Innovative Research in Science, Engineering and Technology*, 3(6), pp. 13755–13762.

Perkins, J. (2014) *Python 3 Text Processing With NLTK 3 Cookbook*, *Python 3 Text Processing With NLTK 3 Cookbook*. Birmingham, UK: Packt Publishing.

Rechenthin, M., Street, W. N. and Srinivasan, P. (2013) 'Stock chatter : Using stock sentiment to predict price direction', *Algorithmic Finance*, 2(3–4), pp. 169–196. doi: 10.3233/AF-13025.

Schumaker, R. P. and Chen, H. (2009) 'Textual analysis of stock market prediction using breaking financial news', *ACM Transactions on Information Systems*, 27(2), pp. 1–19. doi: 10.1145/1462198.1462204.

Sedik, T. S. and Williams, O. H. (2011) *Global and Regional Spillovers to GCC Equity Markets*, International Monetary Fund.

Si, J., Mukherjee, A., Liu, B., Li, Q., Li, H. and Deng, X. (2013) 'Exploiting Topic based Twitter

Sentiment for Stock Prediction', in *Proceedings of the 51st Annual Meeting of the Association for Computational Linguistics (Volume 2: Short Papers)*. Sofia, Bulgaria: Association for Computational Linguistics (ACL), pp. 24–29.

Sornette, D. (2003) 'Why stock market crash', *Vasa*, 32(1), pp. 54–55. doi: 10.1024/0301-1526.32.1.54.

Sulaiman, F., Zailani, S. and Ramayah, T. (2012) 'Intranet Portal Utilization: Monitoring Tool for Productivity - Quality and Acceptance Point of View', *Procedia - Social and Behavioral Sciences*. The Authors, 65, pp. 381–386. doi: 10.1016/j.sbspro.2012.11.138.

Tabak, B. M., Takami, M. Y., Cajueiro, D. O. and Petitinga, A. (2009) 'Quantifying price fluctuations in the Brazilian stock market', *Physica A: Statistical Mechanics and its Applications*. Elsevier B.V., 388(1), pp. 59–62. doi: 10.1016/j.physa.2008.09.028.

Tim and Berners-Lee (2006) 'The Semantic Web Revisited', *IEEE Intelligent Systems*, 21(3), pp. 96–101. doi: 10.1109/MIS.2006.62.

Trigg, R. H. and Weiser, M. (1986) 'TEXTNET: a network-based approach to text handling', *ACM Transactions on Information Systems*. New York, NY, USA: ACM, 4(1), pp. 1–23. doi: 10.1145/5401.5402.

Wasserman, S. and Faust, K. (1994) *Social Network Analysis: Methods and Applications*. New York, NY: Cambridge University Press. doi: 10.1525/ae.1997.24.1.219.

Wenyue, S., Chuan, T. and Guang, Y. (2015) *Network Analysis of the stock market*. CS224W Project Report, Standford University.

Wright, R. W. and Dana, L. P. (2003) 'Changing Paradigms of International Entrepreneurship Strategy', *Journal of International Entrepreneurship*, 1(1), pp. 135–152. doi: 10.1023/A:1023384808859.

Xie, B., Passonneau, R. J. and Wu, L. (2013) 'Semantic Frames to Predict Stock Price Movement', in *Proocedings of the 51st Annual Meeting of ACL*. Sofia, Bulgaria, pp. 873–883.

Yang, S. Y., Mo, S. Y. K. and Liu, A. (2015) 'Twitter financial community sentiment and its predictive relationship to stock market movement', *Quantitative Finance*, 15(10), pp. 1637–1656. doi: 10.1080/14697688.2015.1071078.

Zhang, X., Fuehres, H. and Gloor, P. A. (2011) 'Predicting Stock Market Indicators Through Twitter "I hope it is not as bad as I fear"', *Procedia - Social and Behavioral Sciences*, 26, pp. 55–62. doi: 10.1016/j.sbspro.2011.10.562.